\documentclass[10pt,twocolumn,letterpaper]{article}

\usepackage[pagenumbers]{cvpr} 
\usepackage{amsmath}
\usepackage{amssymb}
\usepackage{amsfonts}
\usepackage{dsfont}
\usepackage{tabularx}
\usepackage{multirow}

\usepackage{algorithm}
\usepackage{algpseudocode}
\usepackage{float}

\usepackage[dvipsnames]{xcolor}

\definecolor{cvprblue}{rgb}{0.21,0.49,0.74}
\usepackage[pagebackref,breaklinks,colorlinks,citecolor=cvprblue]{hyperref}

\title{NoiseCLR: A Contrastive Learning Approach for Unsupervised Discovery of Interpretable Directions in Diffusion Models}

\author{Yusuf Dalva \qquad
Pinar Yanardag \\
Virginia Tech\\
{\tt\small \{ydalva, pinary\}@vt.edu} \\
\small{Project webpage: \url{https://noiseclr.github.io}}
}

\begin{document}
\twocolumn[{
\maketitle
\begin{center}
    \captionsetup{type=figure}
    \vspace{-1em}
\newcommand{\imwidth}{1.0\textwidth}

\begin{tabular}{@{}c@{}}
 
\parbox{\imwidth}{\includegraphics[width=\imwidth, ]{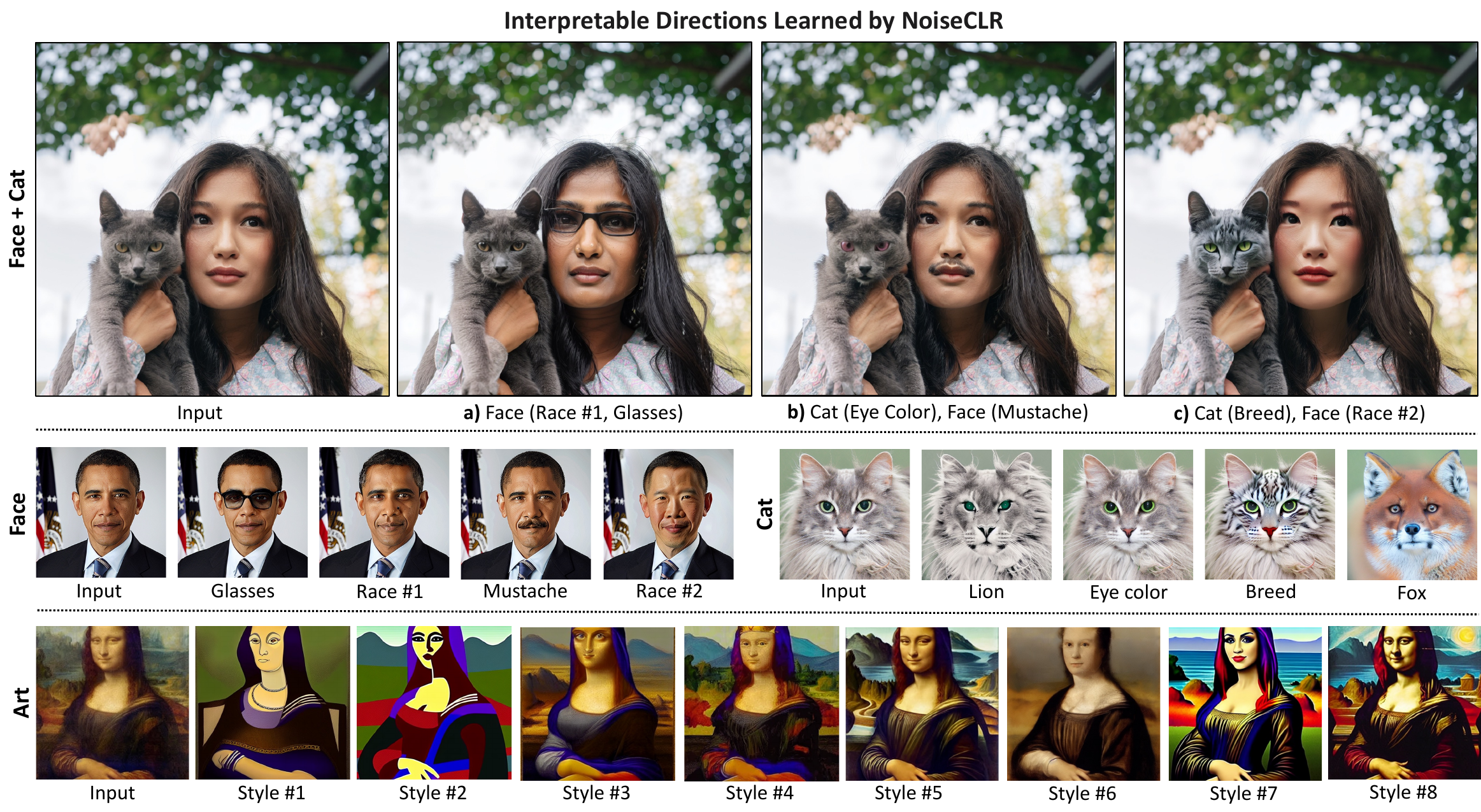}}
\\

\vspace{1em}
\end{tabular}
    \vspace{-2.5em}
    \captionof{figure}{\textbf{NoiseCLR.} We propose an unsupervised approach to identify interpretable directions in text-to-image diffusion models, such as Stable Diffusion \cite{rombach2022high}. Our method finds semantically meaningful directions across various domains like \textit{faces, cats}, and \textit{art}. NoiseCLR can apply multiple directions either within a single domain (a)  or across different domains   in  the same image (b, c) in a disentangled manner. Since the directions learned by our model are highly disentangled, there is no need for semantic masks or user-provided guidance to prevent edits in different domains from influencing each other. Additionally, our method does not require fine-tuning or retraining of the diffusion model, nor does it need any labeled data to learn directions.  \textit{Note that our method does not require any text prompts, the direction  names above are provided by us for easy understanding.} 
} 
    \label{fig:teaser}
\end{center}
}]

\maketitle
\begin{abstract}
 Generative models have been very popular in the recent years for their image generation capabilities.  GAN-based models are highly regarded for their disentangled latent space, which is a key feature contributing to their success in controlled image editing.  On the other hand, diffusion models have emerged as powerful tools for generating high-quality images. However, the latent space of diffusion models is not as thoroughly explored or understood. Existing methods that aim to explore the latent space of diffusion models usually relies on text prompts to pinpoint specific semantics. However, this approach may be restrictive in areas such as art, fashion, or specialized fields like medicine, where suitable text prompts might not be available or easy to conceive thus limiting the scope of existing work. In this paper, we propose an unsupervised method to discover latent semantics in text-to-image diffusion models without relying on text prompts. Our method takes a small set of unlabeled images from specific domains, such as faces or cats, and a pre-trained diffusion model, and discovers diverse semantics in unsupervised fashion using a contrastive learning objective. Moreover, the learned directions can be applied simultaneously, either within the same domain (such as various types of facial edits) or across different domains (such as applying cat and face edits within the same image) without interfering with each other. Our extensive experiments show that our method achieves highly disentangled edits, outperforming existing approaches in both diffusion-based and GAN-based latent space editing methods.
\end{abstract}    
\section{Introduction}
\label{sec:intro}

Denoising Diffusion Models (DDMs) \cite{ho2020denoising} and Latent Diffusion Models (LDMs) \cite{rombach2022high} have received considerable attention for their ability in generating high-quality, high-resolution images across a variety of domains. They have achieved remarkable outcomes in the field of generative modeling, particularly with text-to-image models like Stable Diffusion \cite{rombach2022high} which inspired researchers to employ them for image editing tasks through text prompts or various conditions such as scribble or segmentation maps \cite{zhang2023adding}. 

A fundamental aspect of image editing in generative models is the disentangled application of semantics, which involves making changes that are semantically significant to specific areas of the image without affecting unintended regions \cite{mathieu2019disentangling, xia2022gan}. Previous research has demonstrated that generative adversarial networks (GANs) are particularly effective at disentangled image editing due to their structured latent space, leading to significant research in both supervised and unsupervised exploration of the latent directions in GANs \cite{yuksel2021latentclr, harkonen2020ganspace, shen2020interfacegan}.  

However, while identifying directions in the latent space of GANs is relatively straightforward, such as using principal component analysis on sampled latent vectors to discover semantically meaningful directions \cite{harkonen2020ganspace}, uncovering directions in diffusion models in an unsupervised manner is more challenging. This difficulty arises from the inherent design of diffusion models, which estimates the forward noise independently of the input and manage a significant number of latent variables over several recursive timesteps, unlike the more direct approach in GAN-based models. Therefore, most of the prior work that provides fine-grained control over the generation process in diffusion-based models focus on simple solutions such as blending latent vectors, model fine-tuning, embedding optimization \cite{avrahami2023blended, hertz2022prompt, brack2023sega}. However, these methods depend on user-provided text prompts to pinpoint specific semantics, e.g. \textit{`A photo of a woman with an eyeglass'}.  This approach can be restrictive in areas such as \textit{art, fashion} where appropriate text prompts might not be straightforward to create, or in specialized fields such as the medical domain, which demand extensive domain knowledge to create appropriate text prompts.  This limitation highlights the significance of discovering directions in the latent space in an unsupervised manner.  

\begin{figure}
    \centering
    \includegraphics[width=\linewidth]{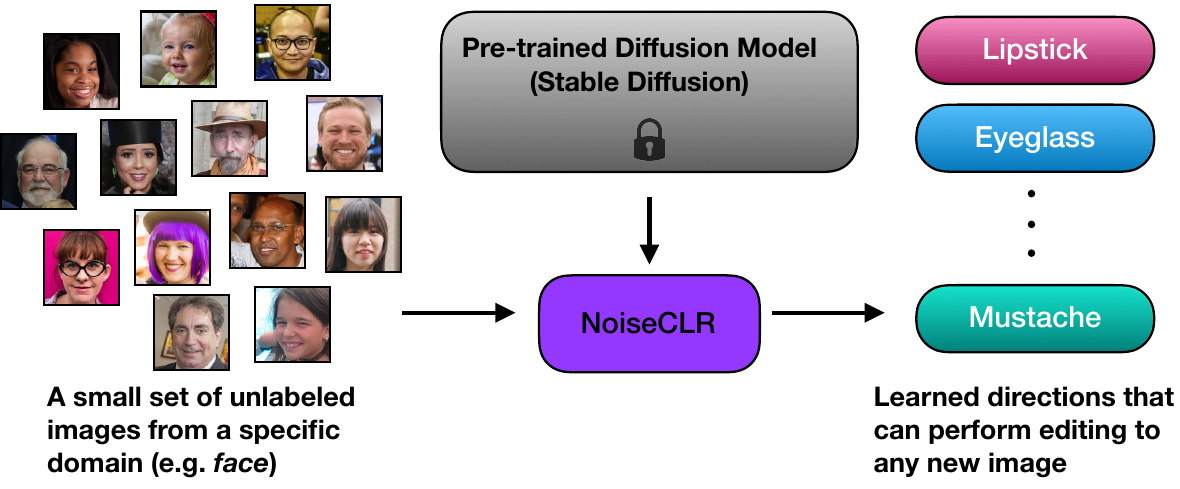}
    \caption{\textbf{NoiseCLR in a nutshell}. Our method employs a pre-trained diffusion model such as Stable Diffusion \cite{rombach2022high}, alongside a small collection of \textit{unlabeled} images from a specific domain such as \textit{faces} or \textit{cats} and learns diverse directions in an unsupervised fashion using a contrastive learning objective. The discovered directions can perform disentangled edits, such as allowing for semantically meaningful edits or adding \textit{lipstick} or \textit{eyeglasses} to any new image. }
    \label{fig:nutshell}
\end{figure}
 
A number of strategies have been introduced to systematically investigate the latent directions within diffusion models \cite{kwon2022diffusion, park2023understanding}. However, much of the existing research  acknowledges limitations when working with large models like Stable Diffusion, often opting for simpler diffusion models such as DDPM \cite{kwon2022diffusion, park2023understanding, haas2023discovering}. These methods fail to fully exploit the capabilities of large-scale models such as Stable Diffusion and rely on separate DDPM models for each domain to identify directions. Therefore, despite significant advancements, a thorough exploration of the latent space in large diffusion-based models like Stable Diffusion remains an ongoing challenge. Discovering directions in latent space of a diffusion-based models is essential not only in the context of image editing but also for a broad spectrum of other applications. First, it allows  more precise control over the image generation process, thereby significantly enhancing the versatility and applicability of the model across a variety of creative and specialized domains.  Second, this approach fosters a more transparent and insightful exploration, demystifying what is often seen as a 'black-box' model, thus making its latent space more understandable. Thirdly, these insights enhance trust and reliability in the model and could be instrumental in identifying and mitigating potential biases, thus fostering further research in the ethical domain.

To the best of our knowledge, our approach is the first unsupervised method that successfully discovers directions in the latent space of Stable Diffusion in a disentangled manner to the extent of combining multiple directions within and across various domains (see Fig. \ref{fig:teaser}). Our contributions are as follows:

\begin{itemize}
    \item We propose NoiseCLR, a contrastive-learning based framework to discover semantic directions in a pre-trained text-to-image diffusion model such as Stable Diffusion. Our approach   does not need textual prompts, labeled data, or user-guidance, relying on a relatively small number of images (around 100) related to the target domain (see Fig. \ref{fig:nutshell}).
    \item Our method demonstrates the ability to discover diverse and fine-grained directions across diverse categories, such as face, cars, cats, and artwork.
    \item Our directions are highly disentangled, can apply multiple directions either within a single domain or across different domains. Our experiments demonstrate that our method can perform edits that are competitive with both state-of-the-art diffusion-based and GAN-based image editing methods.

\end{itemize}

\section{Related Work}
\label{sec:related_work}

\paragraph{Latent Space Exploration of GANs.}
Various techniques have emerged that harness the latent space of GANs for image manipulation \cite{dalva2022vecgan, pehlivan2023styleres, dalva2023image}. Supervised methods often leverage pre-trained attribute classifiers to guide the optimization, facilitating the discovery of meaningful directions within the latent space. Alternatively, they employ labeled data to cultivate classifiers aiming directly at learning desired directions \cite{goetschalckx2019ganalyze,shen2020interfacegan}. Conversely, some studies have demonstrated the potential to identify semantically meaningful directions within the latent space without supervision \cite{voynov2020unsupervised,jahanian2019steerability,upchurch2017deep, yuksel2021latentclr, shen2020closed}. More recent work on GAN-based latent space explorations are pivoting towards utilizing image-text alignment methods such as StyleCLIP \cite{patashnik2021styleclip}.

\paragraph{Latent Space Exploration of Diffusion Models.}
As diffusion-based image generation models are able to synthesize images from various domains, they encode semantically rich content in the form of a latent representations. In order to benefit from these representations, studies attempted to make use of the semantics encoded in the latent space. As a natural extension of latent space discovery, some works \cite{kwon2022diffusion, wu2023latent} attempted to apply image editing by modifying the backward diffusion path using representations learned from the latent variables. While \cite{kwon2022diffusion} formulate their transformation relying on the features learned by the bottleneck block of the denoising model,  \cite{wu2023latent} apply modulation to the latent variables for a target domain, using stochastic diffusion models. In more recent efforts, \cite{park2023understanding} offered a framework to discover latent-specific directions encoding different semantics, inspired by latent space discovery literature in GANs. Even though their approach succeeds in discovering directions in single-domain diffusion models such as DDPMs, their proposed method fails in large-scale diffusion models, such as Stable Diffusion. Moreover, \cite{Liu_2023_ICCV}  decomposes images into a set of composable energy functions representing concepts such as \textit{lighting} or \textit{camera position}.  However, their approach is particularly limited in terms of the number of concepts that can be learned due to memory constraints of their method, and they can only learn conceptualized representations rather than fine-grained latent directions.

\begin{figure*}[!t]
    \centering
    \includegraphics[width=0.95\linewidth]{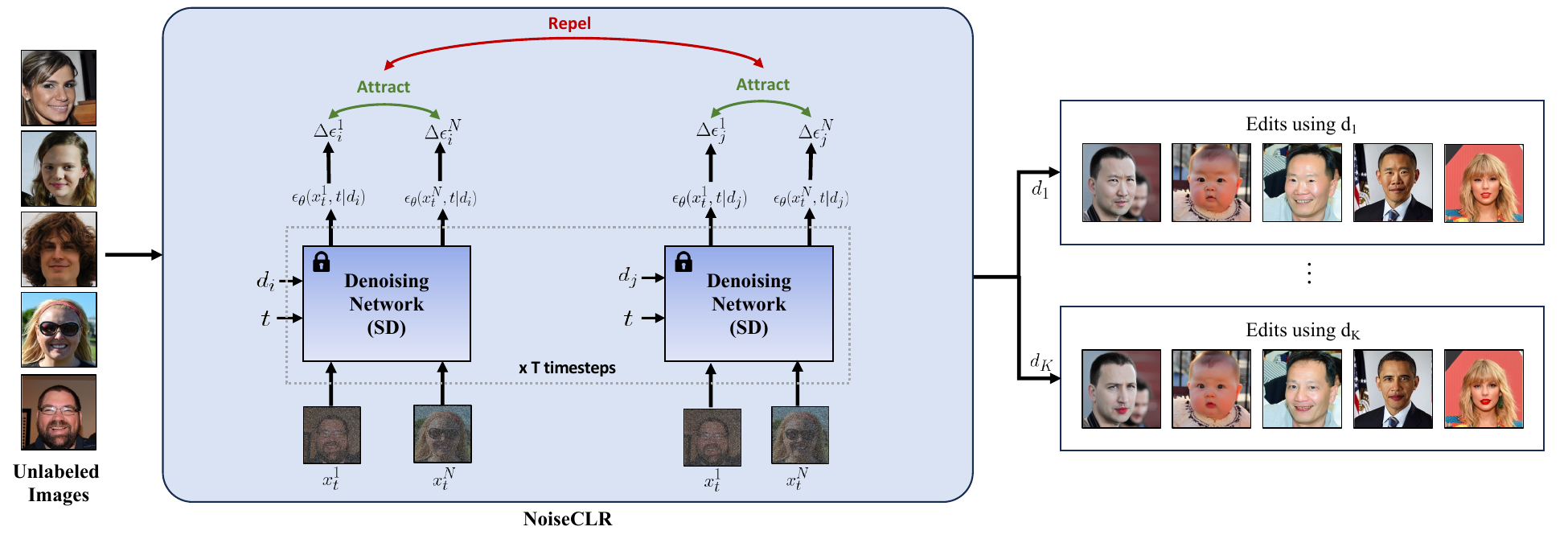}
    \caption{\textbf{NoiseCLR Framework.} NoiseCLR employs a contrastive objective to learn latent directions in an unsupervised manner. Our method utilizes the insight that similar edits in the noise space should attract to each other, whereas edits made by different directions should be repelled from each other. Given $N$ unlabeled images from an particular domain such as facial images, we first apply the forward diffusion process for $t$ timesteps. Then, by using the noised variables $\{x_1, ..., x_N\}$, we  apply the denoising step, conditioning this step with the learned latent directions. Our method discovers $K$ latent directions $d_1, \ldots, d_K$ for a pretrained denoising network such as Stable Diffusion, where directions correspond to semantically meaningful edits such as  \textit{adding a lipstick}. }
    \label{fig:method-overview}
\end{figure*}

\paragraph{Image Editing with Diffusion Models.}
The field of image generation has seen a growing interest towards utilizing diffusion models for editing tasks. One common approach involves supplying text prompts  describing the intended edit. Yet, many implementations result in entangled edits, where unintended sections of the image are altered alongside the target area. Exceptions to this trend can be seen in works like \cite{hertz2022prompt, zhang2023adding} which demonstrate more precise control over the editing process. For instance, ControlNet \cite{zhang2023adding} leverages conditional diffusion model to allow users to manipulate specific image attributes by providing conditions. Likewise, \cite{unitune} allows to perform content-preserving edits by overfitting the diffusion model to the input image. Additionally, \cite{mokady2023null, han2023improving, wu2023uncovering} offer faithful reconstruction of the input image, which makes content-preserving edits with classifier-free guidance possible. Despite being able to preserve the input image while editing, such methods require per-image optimization, which is a bottleneck against real-time image editing. In recent efforts, \cite{wu2023latent} attempts to perform the image editing task by modifying the denoising process of a stochastic diffusion model for the real-editing task. Even though such approaches promise realistic image editing, constructing the ideal prompt for editing is a bottleneck against achieving realistic edits while staying faithful to the original image. To address the flexibility problem, \cite{brack2023sega, liu2022compositional} proposed to compose the desired edit into multiple counterparts. However, these methods face difficulties when applying multiple edits, resulting in entangled results when several changes are made to the same image.

\paragraph{Contrastive Learning.} Contrastive learning has gained traction recently, achieving state-of-the-art results in various unsupervised representation learning tasks. Its principle lies in learning representations by contrasting positive pairs against negative ones \cite{hadsell2006dimensionality}. This approach has found applications in numerous computer vision tasks, including data augmentation \cite{chen2020simple}, diverse scene generation \cite{tian2019contrastive}, random cropping, and flipping \cite{oord2018representation}. In the context of diffusion models, approaches structured with a contrastive setup enabled tasks such as style transfer \cite{Yang_2023_ICCV}, and representation learning \cite{tian2023stablerep}. LatentCLR \cite{yuksel2021latentclr} introduces a method of contrastive learning to identify latent directions in GAN-based models by exploring feature divergences within an intermediate representation of latent vectors. While sharing similar intuitions in terms of utilizing contrastive learning for direction discovery, our approach diverges from LatentCLR. Unlike LatentCLR, which operates on latent vectors sampled from the GAN model, we focus on noise estimations, spanning across multiple diffusion steps. Notably, discovering directions in text-to-image diffusion models is considerably more complex than in GANs. This complexity arises because diffusion models independently estimate forward noise, irrespective of the input, and maintains a significant amount of latent variables across several recursive timesteps.

\begin{figure*}
    \centering
    \includegraphics[width=1\linewidth]{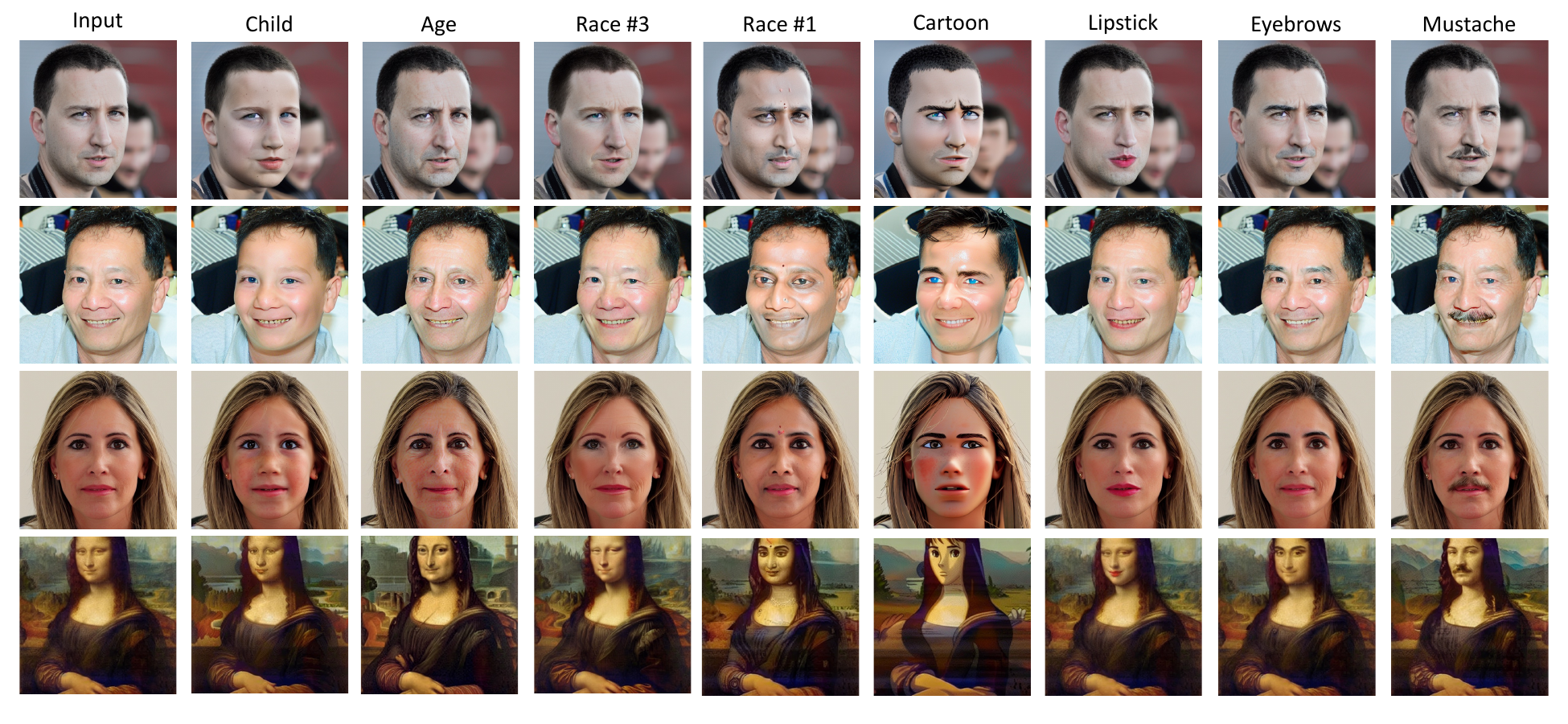}
    \caption{\textbf{Directions learned by NoiseCLR on face domain.} Edits are performed using the directions learned by our method in an unsupervised manner. We annotate the discovered directions above for the sake of understandability. The edits learned by our method are both effective in domain examples (e.g. human faces) and out-of-domain images (e.g. paintings).}
    \label{fig:main_results}
\end{figure*}

\section{Method}
\label{sec:method}

In this section, we describe our proposed method, NoiseCLR, on discovering interpretable directions.  First, we briefly discuss  background  on denoising probabilistic diffusion models. 

\subsection{Denoising Probabilistic Diffusion Models}
Diffusion models \cite{ho2020denoising, song2020denoising, rombach2022high} are generative models that produce data samples through an iterative denoising process, which is often referred as the reverse process. The reverse process involves a set of noise levels $ t \in \{1, ..., T\}$, $\epsilon^t = \alpha^t \epsilon$, where $\epsilon \sim \mathcal{N}(0, 1)$. The denoising network, $\epsilon_\theta$, is designed to estimate the noisy component $\epsilon$ from noised image $x_t$ during the reverse process where $x_t$ refers to the noised version of the real image $x_0$ with a noise level of $\epsilon^t$. The objective function for training such a denoising network is formulated as shown as:

\begin{equation}
    \label{eqn:denoising_network}
    \mathcal{L}_{DM} = \mathbb{E}_{x_0, \epsilon^t \sim \mathcal{N}(0,1), t} \Big[ ||\epsilon^t - \epsilon_\theta(x_t, t)||^2_2 \Big]
\end{equation}

\noindent To generate an image using the denoising network $\epsilon_\theta$, the reverse process is initiated with input $x_T \sim \mathcal{N}(0,1)$. Throughout the reverse diffusion process, the variable $x_t$ is iteratively denoised to get $x_0$ where $t \in \{1, ..., T\}$. The iterative denoising process is formulated as Equation \ref{eqn:reverse_process} for step size $\gamma$ and timestep $t$.
\begin{equation}
    \label{eqn:reverse_process}
    x_{t - 1} = x_{t} - \gamma \epsilon_\theta(x_t, t) + \xi, \; \xi \sim \mathcal{N}(0, \sigma_t^2I)
\end{equation}

\noindent Classifier-free guidance \cite{ho2022classifier} offers a way for conditioned sampling through subtle adjustments in both forward and backward diffusion processes with a specified condition $c$. By training $\epsilon_\theta$ compatible with classifier-free guidance, conditional image generation becomes possible by modifying the noise prediction $\epsilon_\theta(x_t)$ with conditional noise prediction, to get $\Tilde{\epsilon_\theta}(x_t, c)$. For simplicity, we use $\epsilon_\theta(x_t)$ instead of $\epsilon_\theta(x_t, t)$ to represent the predicted noise for timestep $t$, as $t$ is implicitly denoted with variable $x_t$. The predicted noise with classifier-free guidance, $\Tilde{\epsilon_\theta}(x_t, c)$, is defined by Equation \ref{eqn:cfg}:
\begin{equation}
    \label{eqn:cfg}
    \Tilde{\epsilon_\theta}(x_t, c) = \epsilon_\theta(x_t, \phi) + \lambda_g (\epsilon_\theta(x_t, c) - \epsilon_\theta(x_t, \phi))
\end{equation}

where $\phi$ is null-text and  $\lambda_g$ is guidance scale.

\subsection{Contrastive Learning Objective}

The primary objective of NoiseCLR is to learn $K$  semantically meaningful directions, $D = \{d_1, \ldots, d_K\}$, given a small set of $N$ images, $X = \{x_1, \ldots, x_N\}$  in diffusion models in an unsupervised manner. The intuition behind NoiseCLR is best explained as defining an objective that encourages the similarity of the edits done by an arbitrary direction, while discouraging the similarity of edits performed by different directions. In other words, we want for edits carried out by the same direction to be attracted towards each other, while edits conducted by different directions to repel one another, in line with the core principles of contrastive learning. To formulate such an objective, we first define the feature divergences $\Delta \epsilon_k^n$ caused by an arbitrary direction $d_k$ on an arbitrary data sample $x_n$ as follows:

\begin{equation}
    \label{eqn:target_features}
    \Delta \epsilon_k^n = \epsilon_\theta(x_t^n, d_k) - \epsilon_\theta(x_t^n, \phi)
\end{equation}

We define the target feature divergences obtained from $d_j$ and a set of data samples $X' \subset X$ as our positive samples, whereas the target feature divergences for sample $x_i \in X'$ and a set of latent directions $D' \subset D - d_j$ are selected as the negative samples. We formulate our contrastive learning objective in Equation \ref{eqn:noiseclr_loss}:

\begin{equation}
    \label{eqn:noiseclr_loss}
    \mathcal{L} = -log \frac{\sum_{a=1}^{|X'|} \sum_{b=1}^{|X'|} \mathbf{1}_{[a \neq b]} \text{exp}(\text{sim}(\Delta \epsilon_j^a, \Delta \epsilon_j^b) / \tau)}{\sum_{a=1}^{|X'|} \sum_{i=1}^{|D'|} \mathbf{1}_{[i \neq j]} \text{exp}(\text{sim}(\Delta \epsilon_j^a, \Delta \epsilon_i^a) / \tau)}
\end{equation}

\noindent To express the semantic similarity between a pair of target feature differences, we use cosine similarity which is formulated as:
\begin{equation}
    \label{eqn:similarity}
    \text{sim}(\Delta \epsilon_j^a, \Delta \epsilon_j^b) = \frac{\Delta \epsilon_j^a \cdot \Delta \epsilon_j^b}{||\Delta \epsilon_j^a|| \; ||\Delta \epsilon_j^b||}
\end{equation}
\begin{figure*}
    \centering
    \includegraphics[width=1\linewidth]{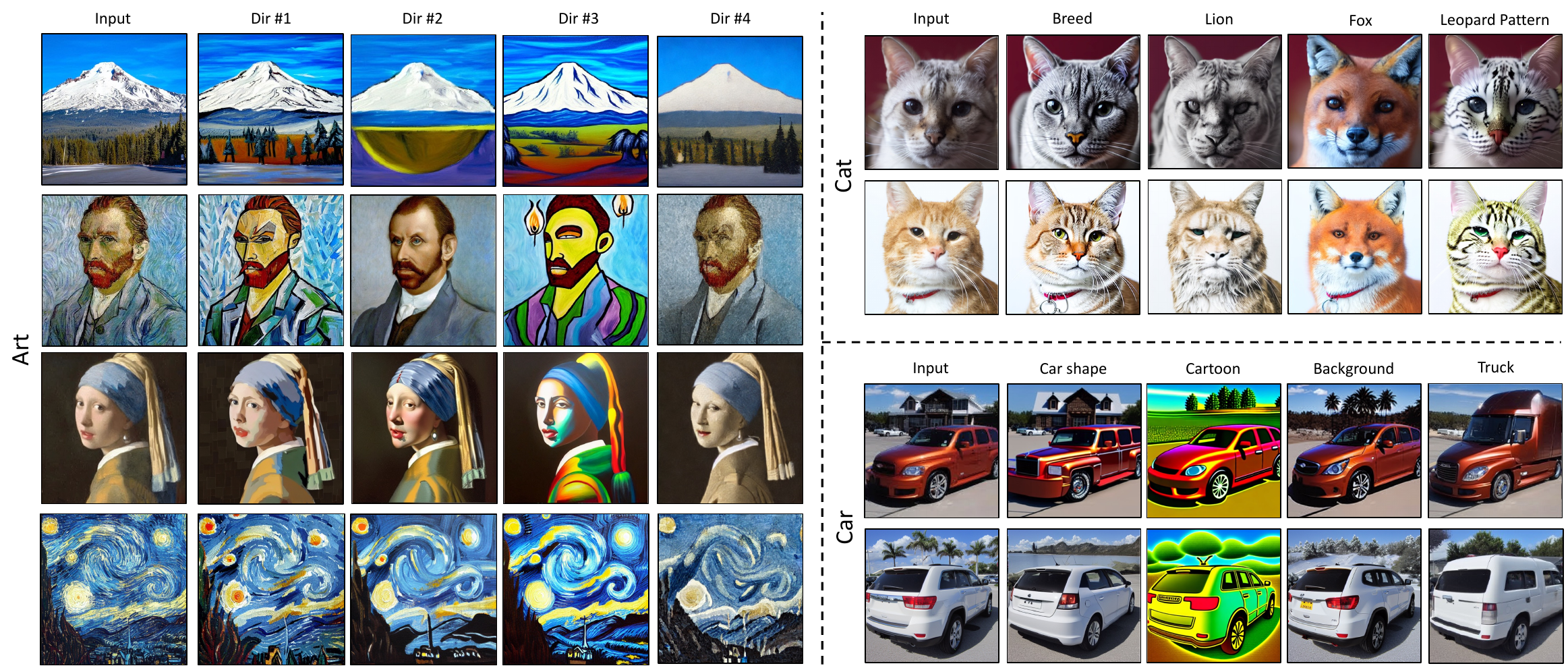}
    \caption{\textbf{Editing results on various domains.} To demonstrate the generalizability of our method across different  domains, we provide editing results on artistic paintings, cats and cars. As demonstrated from in the editing results, our method is able to learn and apply latent directions from various domains using a single diffusion model.}
    \label{fig:multi_domain}
\end{figure*}
\paragraph{Image Editing.}
Given the set of discovered directions $\{d_1, \ldots, d_K\}$, our  editing scheme aims to reflect these semantics to input images in a disentangled manner. To perform such edits, we slightly modify Equation \ref{eqn:cfg} with an editing direction $d_e$ to obtain $\Bar{\epsilon_\theta}(x_t, c, d_e)$ as formulated in Equation \ref{eqn:edit_cfg}, where $c$ serves as the condition used to generate the original image. Leveraging the observation that the difference $\epsilon_\theta(x_t,c) - \epsilon_\theta(x_t, \phi)$ encodes semantic information encoded by the condition $c$, we expand the noise prediction with the difference $\epsilon_\theta(x_t, d_e) - \epsilon_\theta(x_t, \phi)$ for the timestep where editing will be performed:

\begin{equation}
    \label{eqn:edit_cfg}
    \Bar{\epsilon_\theta}(x_t, c, d_e) = \Tilde{\epsilon_\theta}(x_t, c) + \lambda_e (\epsilon_\theta(x_t, d_e) - \epsilon_\theta(x_t, \phi))
\end{equation}

\noindent where $\lambda_e$ denotes the editing scale.

\paragraph{Editing in Multiple Directions.} Since we formulate editing for an arbitrary direction as a summation of predicted noises for a given timestep $t$, we are able to perform multiple edits to a given input variable $x_t$. To perform a set $L$ of discovered directions $\{d_1, \ldots, d_L\}$, we formulate the editing term as a sum of noise predictions, $\Bar{\epsilon_\theta}(x_t, L)$, which is formulated as:
\begin{equation}
    \label{eqn:multi_edit}
    \Bar{\epsilon_\theta}(x_t, L) = \sum_{i=1}^{|L|} \lambda_i (\epsilon_\theta(x_t, d_i) - \epsilon_\theta(x_t, \phi))
\end{equation}
 
\noindent  Using $\Bar{\epsilon_\theta}(x_t, L)$, the overall noise prediction for timestep $t$ is formulated as $\Bar{\epsilon_\theta}(x_t, c, L) = \Tilde{\epsilon_\theta}(x_t, c) + \Bar{\epsilon_\theta}(x_t, L)$.

\paragraph{Real Image Editing.} In addition to performing edits on generated images, we expand our editing approach such that the discovered edits are applicable to real images. Different than sampling fake images from $x_T \sim \mathcal{N}(0,1)$, we initially apply DDIM Inversion \cite{mokady2023null} to obtain this initial variable $x_T$. Using this inverted variable, we reformulate $\Bar{\epsilon_\theta}(x_t, c, d_e)$ as $\Bar{\epsilon_\theta}(x_t, d_e)$ since the image is conditioned by the initial variable $x_T$ only. The formulation for real image editing with a single direction is provided as  follows:
\begin{equation}
    \label{eqn:real_edit}
    \Bar{\epsilon_\theta}(x_t, d_e) = \epsilon_\theta(x_t, \phi) + \lambda_e (\epsilon_\theta(x_t, d_e) - \epsilon_\theta(x_t, \phi))
\end{equation}
\noindent Note that, our approach in editing with multiple directions is also applicable for real images.

\section{Experiments}
\label{sec:experiments}

To assess the effectiveness of NoiseCLR in identifying semantically meaningful latent directions and demonstrate the generalizability of our method, we conducted evaluations across various domains, including human faces, cats, cars, and paintings.

\paragraph{Experimental Setup}  We used  Stable Diffusion-v1.5  for all of our experiments. We used several diverse datasets including FFHQ \cite{karras2019style}, AFHQ-Cats \cite{choi2020starganv2}, and Stanford Cars datasets \cite{KrauseStarkDengFei-Fei_3DRR2013}. For the artistic domain, we perform our experiments on a small subset of paintings to discover latent directions corresponding to artistic styles. In our default setting, we train NoiseCLR with $N = 100$, $K = 100$ and $\tau = 0.5$. To optimize the directions, we use a learning rate of $10^{-3}$ and batch size of 6 for AdamW optimizer \cite{loshchilov2017decoupled}.   Throughout these experiments, we train our directions with relatively modest dataset sizes such as 100 for each domain.  For face and painting domains, we set the size the set of directions to be learned as $|D| = 100$ and for the domains of cats and cars, we set the number of directions to be $|D| = 50$. In each experiment, we set the size of the subset of directions to be used $|D'|$ at every iteration as 20. To ensure the reproducibility of our experiments, we conduct all experiments with a fixed random seed of 0.  Moreover, training our method on a single domain requires approximately 7 hours to learn 100 directions in face domain, and once trained, performing any edit in a zero-shot manner takes about 5 seconds using a single NVIDIA L40 GPU. 

\noindent \textbf{Ablation studies} Prior work such as \cite{wu2023uncovering} and \cite{hertz2022prompt} has shown that timesteps are crucial factors affecting the disentanglement capability of Stable Diffusion. Our method modifies the noise prediction for a certain interval of timesteps to achieve more disentangled edits. As a rule of thumb, we apply the discovered edits starting from $t = 0.5T$. However, for edits that require changes in the coarse structure of the input (e.g. eyeglasses), editing at earlier timesteps are required (within the interval [$0.9T, 0.8T$]). Moreover, our method has the capability to learn from randomly generated images, as opposed to real images such as FFHQ. Nevertheless, our  experiments have shown that  randomly generated images in SD frequently contain artifacts and biases. Additionally, our method is capable of learning directions from as few as 10 images in a small subset. However, our findings indicate that generally, using 100 real images  leads to faster convergence. To explore in-depth ablation studies concerning timesteps, the number of images, the number of directions, and the utilization of real/fake images during training, please refer to \ref{sec:ablation}.

 \begin{figure}[!ht]
   \centering
    \includegraphics[width=1\linewidth]{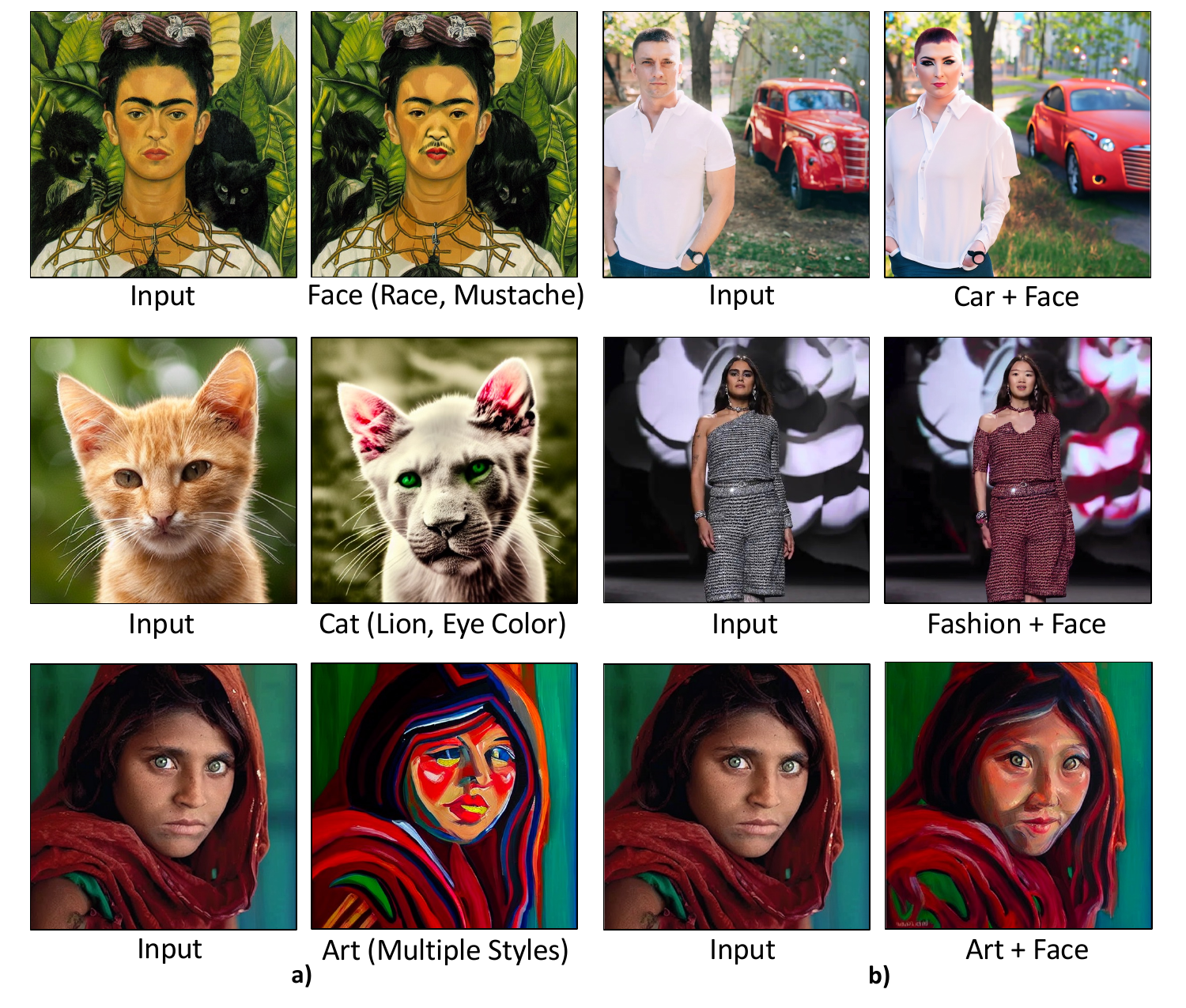}
    \caption{\textbf{Intra-domain and Cross-domain Editing.} Our method can find  domain-specific edits that can be composed either a) intra-domain where edits from the same domain can be applied simultaneously, b) cross-domain, where edits from different domains can be combined and applied simultaneously. }
    \label{fig:compositional}
\end{figure}

\subsection{Qualitative Results}
Unlike existing methods for exploring the latent space of diffusion models, our approach can identify latent directions of various domains using a single diffusion model. Since face images carry significant amount of variance in terms of facial features and is one of the most popular type of edits in both GAN and diffusion-based models, we first investigate the face editing capabilities of the directions discovered by NoiseCLR. Fig. \ref{fig:main_results} displays a variety of distinct directions, including broad edits that can change the overall structure of the face, like \textit{aging} or \textit{race}, as well as more detailed directions that modify fine-grained facial features, such as \textit{lipstick} or a \textit{mustache}. We emphasize that our edits are highly disentangled, meaning they achieve the intended modification without affecting any unintended parts. Given that our method conducts direction discovery in an entirely unsupervised manner, it has the freedom to explore the semantic space of the diffusion model during training. Consequently, NoiseCLR can discover directions not explicitly represented in the input dataset but still compatible with the domain of the training images, such as  \textit{cartoon} direction. Additionally, as NoiseCLR leverages the semantic understanding of the diffusion model, face edits generalize well to artistic paintings, as demonstrated in the last row of Fig. \ref{fig:main_results}. Besides the facial domain, we also showcase the efficacy of our approach in the domains of art, cats, and cars. Qualitative results   are presented in Fig. \ref{fig:multi_domain}. As evident from the figure, NoiseCLR is capable of learning diverse semantics across various domains. This includes identifying multiple \textit{artistic} styles, directions for transforming cats into foxes or lions, and directions for converting cars into trucks.

\begin{figure}[!t]
    \centering
    \includegraphics[width=1\linewidth]{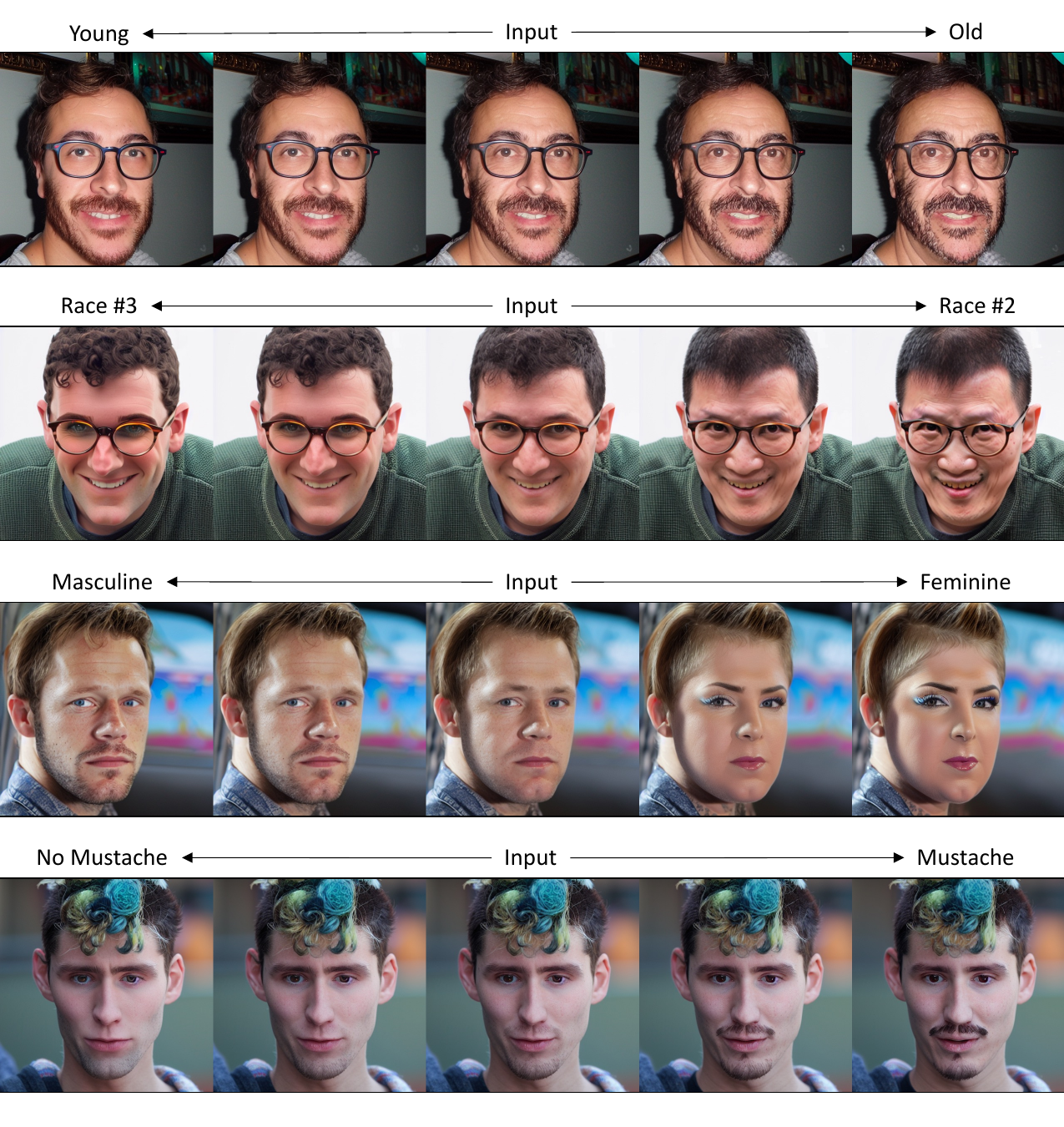}
    \caption{\textbf{NoiseCLR Interpolation Results.} Our method allow users to control  the editing effect using a scale parameter. This feature allows users to either diminish or amplify the effect of the editing direction. For example, with the 'Age' direction, users can reduce the aging effect (Young) or increase it when applied with a positive scale (Old).}
    \label{fig:interpolation}
\end{figure}

\begin{figure*}[!h]
    \centering
    \includegraphics[width=1\linewidth]{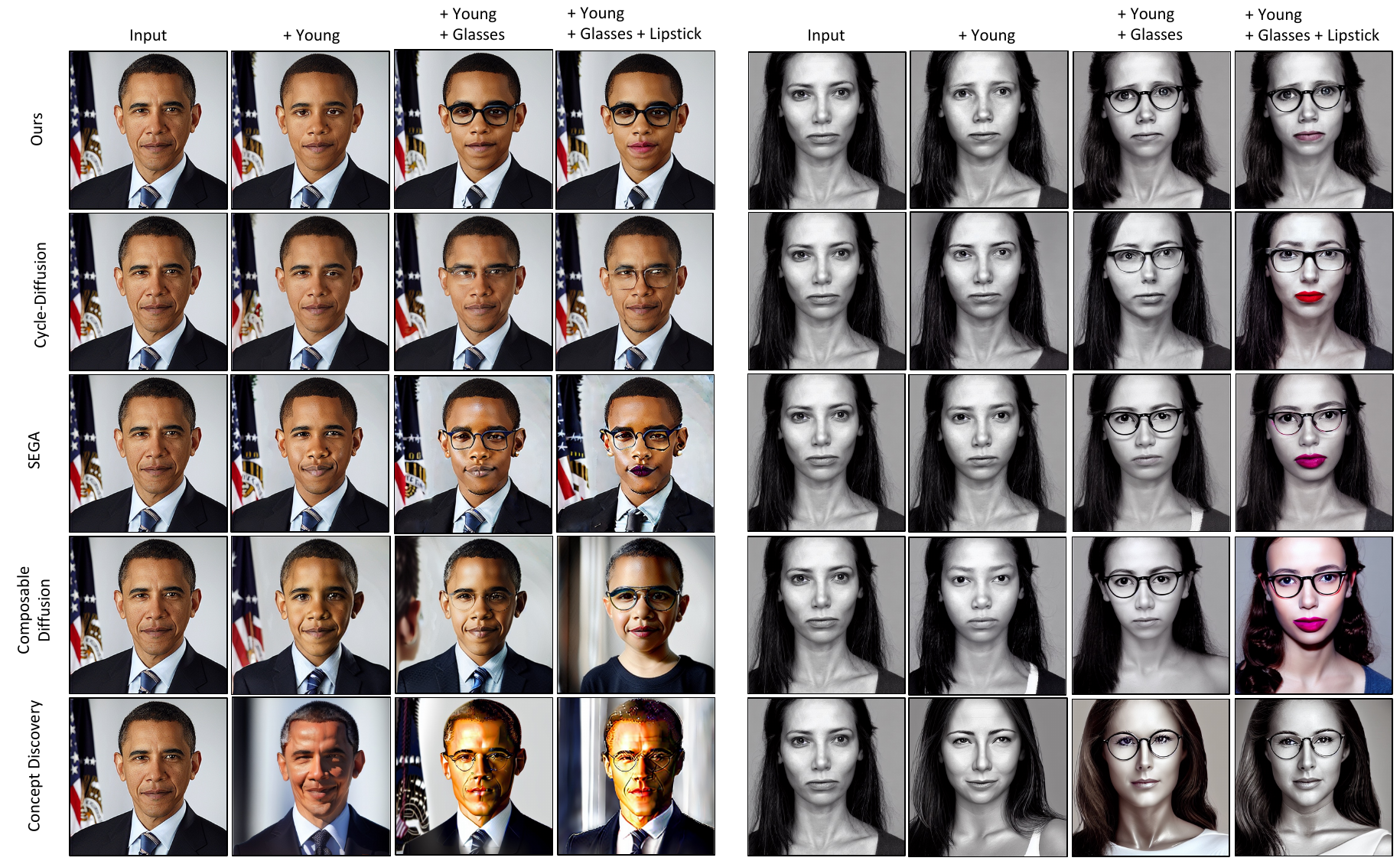}
    \caption{\textbf{Qualitative Comparisons.} We compare our method with Cycle-Diffusion\cite{wu2023latent}, SEGA \cite{brack2023sega} and Composable Diffusion \cite{liu2022compositional} in terms of image editing capabilities and Unsupervised Concept Discovery \cite{Liu_2023_ICCV} to assess the quality of the representations learned by NoiseCLR. We present our comparisons for both real-image editing task and conditional image generation with the provided semantics. As it can be observed from the presented qualitative results, NoiseCLR succeeds over competing methods both in terms of disentangled image editing and learning fine-grained latent directions. }
    \label{fig:quali_comp}
\end{figure*}
\noindent \textbf{Composing Edits} Since NoiseCLR can learn latent directions from different domains within the shared latent space of Stable Diffusion, it is capable of executing both intra-domain and inter-domain edits (see Fig. \ref{fig:teaser} and Fig. \ref{fig:compositional}):

\begin{enumerate}
    \item \textbf{Within the same domain}, where multiple face edits can be applied simultaneously to a single image, as depicted in Fig. \ref{fig:teaser} and Fig. \ref{fig:compositional} (a). Using the same Stable Diffusion model, edits in the face domain can be simultaneously applied to a single image, allowing for changes like altering \textit{race} and adding a \textit{mustache}, as illustrated in Fig. \ref{fig:compositional} (a), top row. Similarly, in the cat domain, our method can concurrently apply edits affecting \textit{eye color} and \textit{lion}, as shown in Fig. \ref{fig:compositional} (a), middle row. Our method can also combine multiple styles in Art domain, as shown in Fig. \ref{fig:compositional} (a), bottom row.

    \item  \textbf{Across different domains}, enabling the application of edits from various domains on the same image simultaneously. For instance, a face edit and a cat edit can be combined together as in Fig. \ref{fig:teaser}. Moreover, our method can apply edits in the \textit{car} domain to transform a car into a \textit{sports car}, while keeping its original color and preserving the \textit{background} (Fig. \ref{fig:compositional} (b), top row). Concurrently, it can alter the \textit{gender} of a person in the image using a face edit. In the same vein, within the fashion domain, our method can change the \textit{color} of a dress, while a face domain edit can modify the \textit{race} of the person wearing the dress (Fig. \ref{fig:compositional} (b), middle row). Our method can also combine face and Art directions, as shown in Fig. \ref{fig:compositional} (b), bottom row.
\end{enumerate}

\noindent \textbf{Interpolating Edits} Our method enables users to modulate the editing effect using a scale parameter. As illustrated in Fig. \ref{fig:interpolation}, it can perform edits along both negative and positive scales. This feature allows users to either diminish or amplify the effect of the editing direction. For example, with the 'Age' direction, users can reduce the aging effect or increase it when applied with a positive scale. Additionally, our method achieves these interpolations in a disentangled manner, ensuring that the edits in both positive and negative directions remain faithful to the original image.

\noindent \textbf{Qualitative comparison} We compare our method with recent approaches, namely, Cycle-Diffusion \cite{wu2023latent}, SEGA \cite{brack2023sega}, Composable Diffusion \cite{liu2022compositional}, and Unsupervised Concept Discovery \cite{Liu_2023_ICCV} methods. Over the compared methods, it is evident that \cite{wu2023latent} and \cite{brack2023sega} struggles with the real image editing task, when multiple semantics are modified even though the edits are performed in a disentangled manner. On the other end, the edits performed by \cite{liu2022compositional} manages to preserve the edit quality, whereas they suffer preserving the image contents. As can be seen in Fig. \ref{fig:quali_comp}, NoiseCLR outperforms the competing approaches both in terms of semantic faithfulness and disentanglement capabilities. For comparisons with diffusion and GAN-based editing methods, please refer to \ref{sec:extra}.

\subsection{Quantitative Results}

 In evaluating the efficacy of discovered edits, we conduct a re-scoring analysis on the directions representing the semantics of "Indian", "Asian", "mustache", "child" and "lipstick" which are arbitrarily selected from the directions discovered by NoiseCLR. In our analysis, we assess the change in classification probability of the CLIP classifier \cite{radford2021learning} in the desired attribute and examine if the directions are disentangled between each other (see Table \ref{tab:rescoring}).  In the presented scores, an increase corresponds to increased classification confidence for the subjected semantic, whereas a decrease implies the decrease of the presence of the semantic. Ideally, we expect the scores of the unedited semantics to change minimally, whereas the confidence for the edited semantic should increase. Relying on the re-scoring analysis performed, we consider our edits disentangled as the semantics that are not naturally related do not change significantly. However, we also acknowledge that applying the edit of "Child" significantly changes the race-based edits ("Indian", "Asian"). We relate this due to internal biases in Stable Diffusion. Moreover, we  compared   LPIPS \cite{zhang2018unreasonable} scores to measure  how well the similarity to the original image distribution is maintained. Table \ref{tab:scores_lpips} shows the results for several edits. The LPIPS metrics clearly demonstrate that our method consistently achieves lower LPIPS scores compared to other approaches, signifying improved coherence during the editing process.

\paragraph{User Study.} We assess the editing capabilities of our model through an user study conducted on 50 participants using Amazon Mturk platform. For each of the methods that we compare with, we show volunteers several edits performed with common semantics and asked to determine if they consider the performed edit successful in terms of the given semantic, and if the edit is performed in a disentangled way. For each question, the users are asked to give a rating between 1-5 to indicate their preference where 5 means the highest score (see Table \ref{tab:user_study}). Our results demonstrate that NoiseCLR attained higher scores in both edit quality and disentanglement evaluations, underscoring the superior performance of our approach.

\begin{table}
    \centering
    \begin{tabular}{@{}lccccc@{}}
        \toprule
         &  Indian &  Asian & Mustache & Child & Lipstick\\ \hline
        Indian & \textbf{29.8} & 16.1 & 8.1 & 5.6 & 4.7\\
        Asian & -10.5 & \textbf{27.5} & 0.0 & -2.0 & 1.2 \\ 
        Mustache & 3.6 & -7.6 & \textbf{48.9} & -13.2 & 1.7 \\ 
        Child & -37.7 & -14.1 & 2.3 & \textbf{32.8} & 11.7\\
        Lipstick & -8.9 & -3.4 & 7.3 & -0.7 & \textbf{11.0} \\\hline
    \end{tabular}
    \caption{\textbf{Re-scoring Analysis.} The change in classification probability of the CLIP classifier for various attributes. Bold numbers indicate that NoiseCLR consistently enhances the target semantics across all attributes. Additionally, our approach achieves disentangled editing by minimizing its influence on other attribute scores when modifying a single attribute. }
    \label{tab:rescoring}
\end{table}

\begin{table}
    \centering
    \begin{tabular}{@{}lcc@{}}
        \toprule
         \textbf{Method} & Edit Quality $\uparrow$ & Disentanglement  $\uparrow$\\ \hline
        Composable D. \cite{liu2022compositional} & 2.19 & 2.15 \\
        Concept D. \cite{Liu_2023_ICCV} & 1.20 & 1.28 \\ 
        SEGA \cite{brack2023sega} & 1.52 & 1.81 \\ 
        Cycle-Diffusion \cite{wu2023latent} & 1.87 & 2.73 \\ \hline
        Ours & \textbf{2.65} &  \textbf{3.05} \\ \hline
    \end{tabular}
    \caption{\textbf{User Study Results.} The average response score of the participants are provided in the table. The scoring is performed within the scale of 1-to-5.}
    \label{tab:user_study}
\end{table}

\begin{table}
    \centering
    \begin{tabular}{l|cccc} \hline
        \textbf{Method} & Age & Mustache & Gender & Race \\ \hline
        Composable D. & 0.19 & 0.40 & 0.42 & 0.40 \\
        Cycle-Diffusion & \textbf{0.10} & 0.21 & 0.23 & 0.26 \\ 
        SEGA & 0.11 & 0.23 & 0.27 & 0.27 \\
        Ours & 0.17 & \textbf{0.17} & \textbf{0.20} & \textbf{0.13} \\ \hline
    \end{tabular}
    \caption{LPIPS \cite{zhang2018unreasonable} scores (lower is the better). Our method is able to achieve lower LPIPS than the other methods, indicating greater coherence while performing the edits.}
    \label{tab:scores_lpips}
\end{table}

\section{Limitations}

Our method is built upon the pre-trained Stable Diffusion model. Consequently, its manipulation capabilities are heavily dependent on the datasets Stable Diffusion was trained on, as well as the language model CLIP utilized by Stable Diffusion. While the joint representation capabilities of CLIP are impressive, they also have limitations and can exhibit biases towards certain attributes.  Furthermore, similar to other image synthesis tools, our framework raises concerns about potential misuse for malicious purposes \cite{korshunov2018deepfakes}.

\section{Conclusion}
\label{sec:conclusion}
We present an approach to discover to discover latent directions in large text-to-image diffusion models in an unsupervised way using a novel contrastive learning framework.  Our method can  combine multiple directions within and across various domains, such as face, cars, cats and artwork. Our experiments demonstrate that our method can perform edits that are competitive with both state-of-the-art diffusion-based and GAN-based image editing methods. Our approach not only provides more precise control over the image generation process, greatly expanding the model's versatility and usability in diverse creative and specialized fields, but it also promotes a more transparent and insightful exploration. This helps to demystify what is often perceived as a 'black-box' model. Furthermore, these insights increase trust and reliability in the model and could play a crucial role in identifying and addressing potential biases, thereby encouraging further research in ethical considerations. 

{
    \small
    \bibliographystyle{ieeenat_fullname}
    \bibliography{main}
}

\clearpage
\makeatletter
\renewcommand \thesection{S.\@arabic\c@section}
\renewcommand\thetable{S.\@arabic\c@table}
\renewcommand \thefigure{S.\@arabic\c@figure}
\makeatother
\setcounter{section}{0}
\setcounter{page}{1}
\maketitlesupplementary

\section{Additional Comparisons}
\label{sec:extra}

\subsection{GAN-based Editing Methods}

\begin{figure*}
    \centering
    \includegraphics[width=0.8\linewidth]{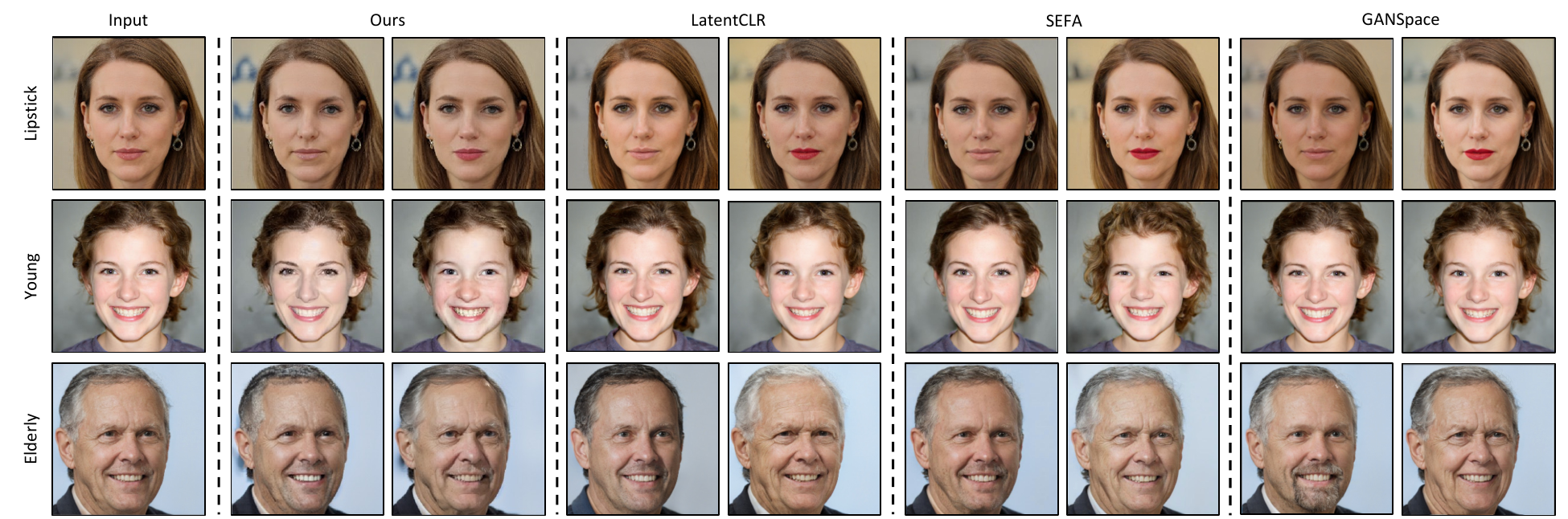}
    \caption{\textbf{Comparisons with GAN-based Latent Discovery Methods.} We also compare NoiseCLR with latent direction discovery methods on GANs. As it is also demonstrated in our results, our editing \& direction discovery method produces competitive results with GAN-based methods, in terms of fine-grained face editing.}
    \label{fig:quali_comp_gan}
\end{figure*}

GAN-based editing methods are known to have superior editing capability due to their disentangled latent space \cite{xia2022gan}. In our analysis, we also include a comparison of NoiseCLR with several state-of-the-art GAN-based methods that find directions in the latent space in an unsupervised manner, LatentCLR \cite{yuksel2021latentclr}, GANspace \cite{harkonen2020ganspace}, and SeFa \cite{shen2021closed} (see Fig. \ref{fig:quali_comp_gan}). As can ben seen from the figure, our diffusion-based model achieves competitive results when compared with its GAN-based counterparts.

\subsection{Diffusion-based Editing Methods}

In this section, we provide additional results across other methods using qualitative and quantitative comparisons. 

\subsection{Qualitative Comparisons}
\label{sec:pullback}

\begin{figure*}[!h]
    \centering
    \includegraphics[width=1\linewidth]{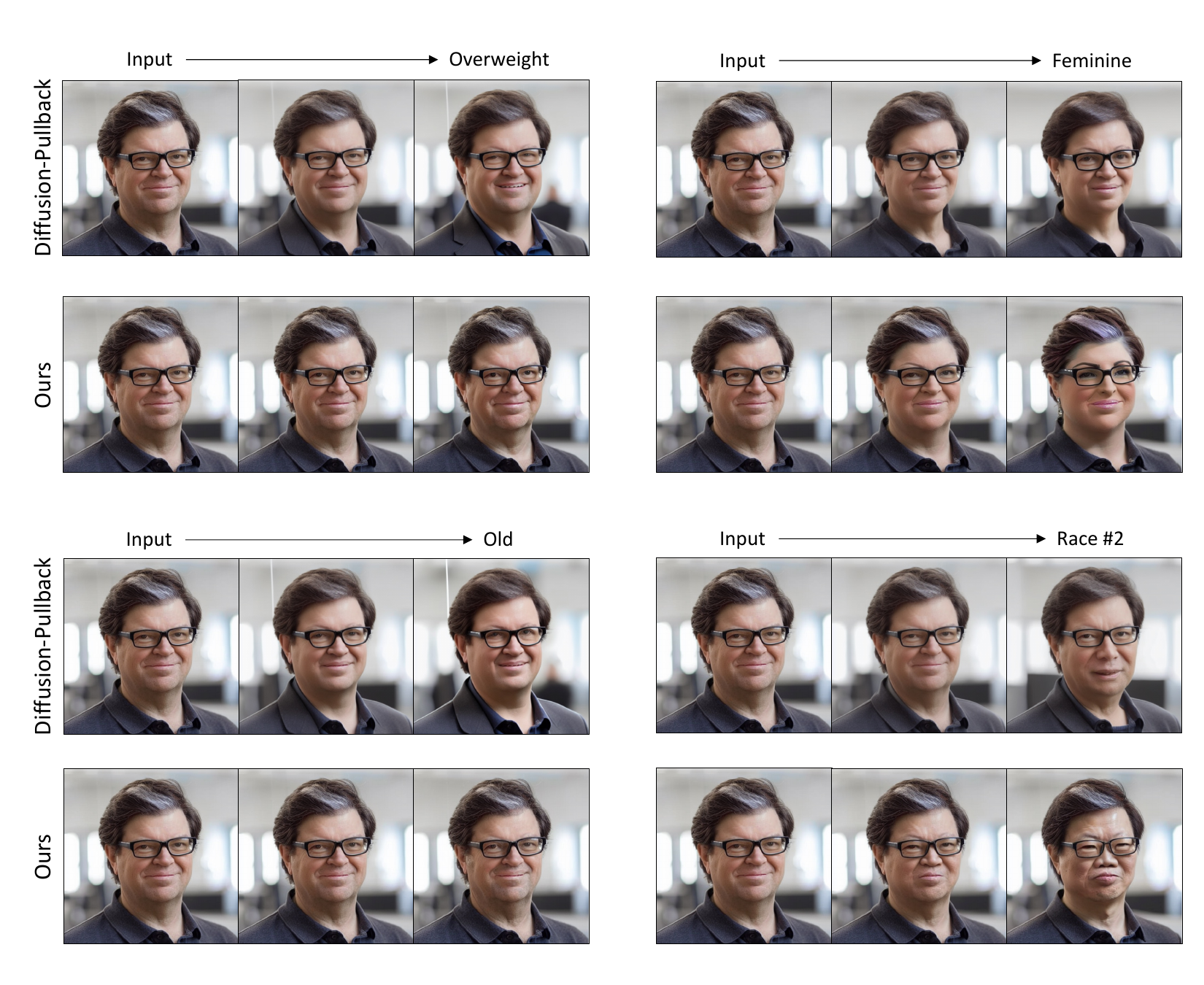}
    \caption{Our comparison with Diffusion-Pullback \cite{park2023understanding} focuses on the \textit{overweight} and \textit{gender} edits, the sole directions provided in \cite{park2023understanding} (utilizing the unsupervised version of their method on Stable Diffusion for face edits). The additional directions, \textit{Old} and \textit{Race}, were identified by ourselves after applying their method to 50 directions. The comparisons clearly demonstrate that our method not only executes edits more faithfully compared to \cite{park2023understanding}, but it also uncovers a significantly greater number of directions, as detailed in the main paper (refer to Fig. \ref{fig:main_results}). }
    \label{fig:nips_comparison}
\end{figure*}

\begin{figure*}
    \centering
    \includegraphics[width=1\linewidth]{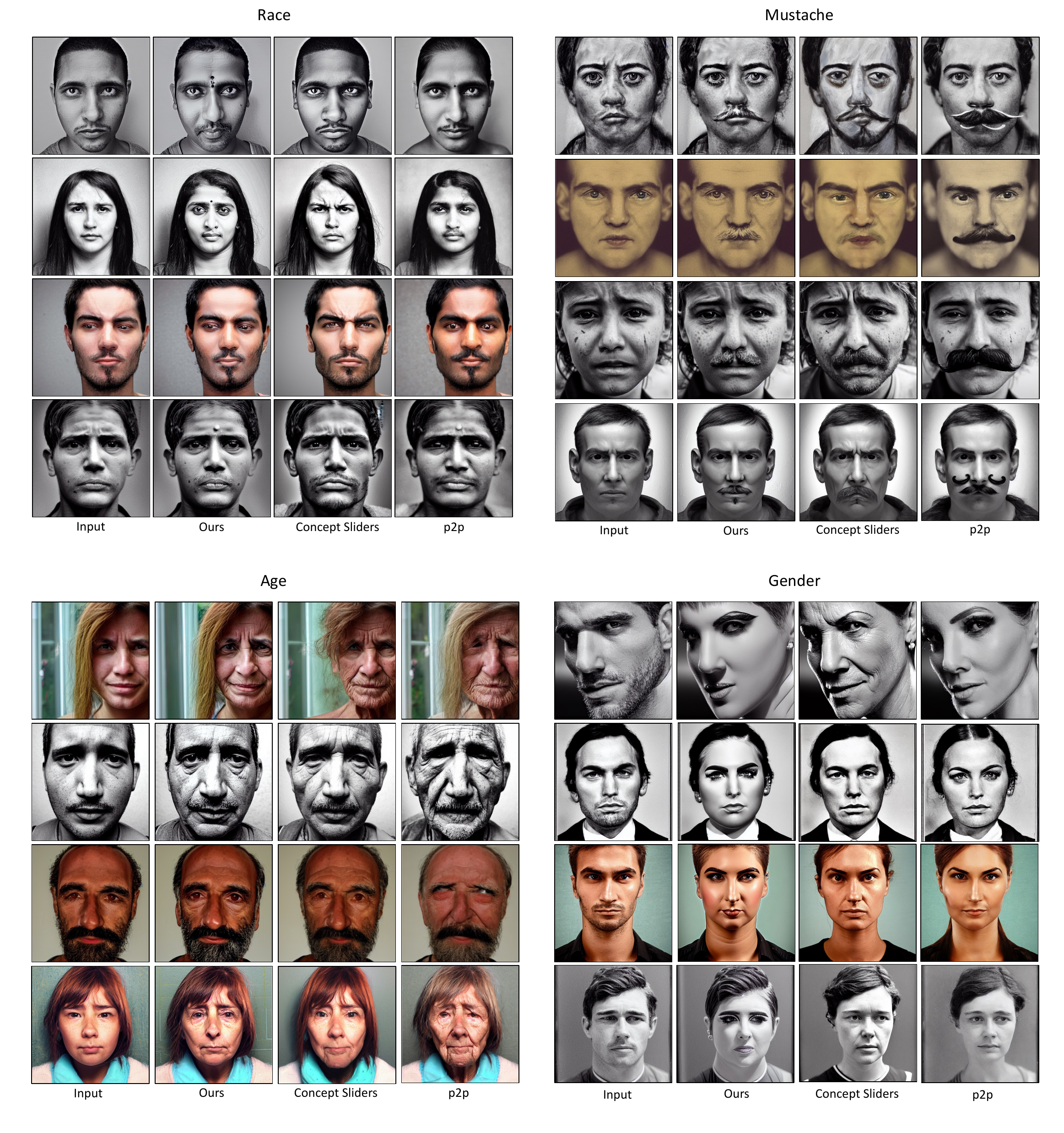}
    \caption{Comparison on Race, Mustache, Age and Gender attributes with our method, Concept Sliders\cite{gandikota2023sliders} and Prompt2Prompt (p2p) \cite{hertz2022prompt}. Although all methods are capable of accomplishing the intended edit to some degree, our method stands out by remaining more faithful to the original input image and ensuring the edits are disentangled. For instance, Prompt2Prompt (P2P) tends to significantly modify the original image, deviating considerably from the initial input. Similarly, Concept Sliders often alter the facial shape (as observed in the Race edit) and mix changes in the face with age attributes in the mustache edit, leading to entangled edits. }
    \label{fig:edit_comp}
\end{figure*}

\begin{itemize}
      \item \textbf{Diffusion-Pullback}  \cite{park2023understanding}  proposes an unsupervised direction discovery method in diffusion-based models. Their approach utilizes the pullback metric to identify latent bases for image editing and optionally incorporates text prompts. While they achieved promising results with DDPM-based models (note that this needs a \textit{separate DDPM model for each domain such as face, cats, and so on}), they report that their application to Stable Diffusion didn't fully realize its potential. Specifically, their method uncovered only a limited number of directions in Stable Diffusion, e.g. only two reported for face edits in their paper. They noted that some of the latent vectors they discovered led to sudden and drastic changes during the editing process. This issue was attributed to the complex geometry of the latent space, which poses a challenge for achieving smooth and seamless edits.

    In their study, only two editing directions called 'overweight' and 'gender' were initially reported. For a fair comparison, we used the same input image from their paper and created edited results for these directions using our method. Additionally, we run their source code to discover two more directions, 'Old' and 'Race', and reported the results. Please refer to Fig. \ref{fig:nips_comparison}. 
    
    The comparisons demonstrate that our method not only executes edits more faithfully compared to \cite{park2023understanding}, but it also uncovers a significantly greater number of directions (refer to Fig. \ref{fig:main_results} in the main paper). 
    
  \item \textbf{Concept sliders} \cite{gandikota2023sliders}, a concurrent work with ours, either rely on text prompts or paired image data for editing images. For instance, to edit \textit{eyebrow shape} of a  face image, one would need a text prompt like `eyebrows' or a pair of images showing the person \textit{before} and \textit{after} their eyebrow shape changes. This reliance on text prompts or paired data for defining edits aligns them with supervised methods in image editing. 
    
    Please see Fig. \ref{fig:edit_comp} for a comparison of their edits (found via providing text prompts defining the edits) vs. ours (found via unsupervised discovery). Although Concept Sliders are capable of accomplishing the intended edit to some degree, our method stands out by remaining more faithful to the original input image and ensuring the edits are disentangled. For instance,  Concept Sliders often alter the facial shape (as observed in the Race edit) and mix changes in the face with aging in the mustache edit, leading to entangled edits.

    \item \textbf{Prompt2Prompt} is an image editing method that uses cross-attention, and requires both a source and target text prompt. We compare our editing results with Prompt2Prompt in Fig. \ref{fig:edit_comp}. Note that since their method does not discover a direction, they are only able to perform a single edit and do not have the ability to control the scale of the edit applied to the image, which limits their usage. On the other hand, the requirement of a source prompt poses another limitation, which might not be feasible in domains such as art. Nevertheless, our results show that our method is able to perform the desired edits in a much more disentangled way while being faithful to the input image.  Notice that Prompt2Prompt (P2P) tends to significantly modify the original image, deviating considerably from the initial input as can be seen from Age edit (Fig. \ref{fig:edit_comp} bottom left) or performs unrealistic edits as in Mustache edit (Fig. \ref{fig:edit_comp} top right).

    \item \textbf{DiffusionDisentanglement} Note that although we intended to compare our results with those from Wu et al. \cite{wu2023uncovering}, an editing method that optimizes weights to perform disentangled edits given a text prompt, we were unable to do so due to the high GPU requirements of their method (namely, 48GB for a single edit). \footnote{\url{https://github.com/UCSB-NLP-Chang/DiffusionDisentanglement/issues/6}.}
\end{itemize}

\begin{table}[ht]
    \centering
    \begin{tabular}{|l|c|c|c|c|}
    \hline
         \textbf{Method} & Race & Mustache & Age & Gender\\ \hline
         Prompt2Prompt \cite{hertz2022prompt} & 0.24 & 0.22 & 0.28 & 0.25 \\ \hline
         Concept Sliders \cite{gandikota2023sliders} & 0.18 & 0.13 & 0.21 & 0.25 \\ \hline
         Ours & \textbf{0.15} & \textbf{0.12} & \textbf{0.18} & \textbf{0.21} \\ \hline
    \end{tabular}
    \caption{LPIPS \cite{zhang2018unreasonable} metric which measures how well the similarity to the original image distribution is maintained (lower is the better). Our method is able to achieve lower LPIPS than the other methods, indicating greater coherence while performing the edits.}
    \label{tab:lpips2}
\end{table}
\subsection{Quantitative Comparisons}
We also compared Concept Sliders \cite{gandikota2023sliders} and Prompt2Prompt \cite{hertz2022prompt} methods in a quantitative way using LPIPS \cite{zhang2018unreasonable} metric which measures how well the similarity to the original image distribution is maintained. Table \ref{tab:lpips2} shows the results for \textit{Race, Mustache, Age, Gender} edits. As can be seen from the LPIPS metrics, our method is able to achieve lower LPIPS than the other methods, indicating greater coherence while performing the edits.

\section{Ablation Study}
\label{sec:ablation}

In this section, we perform ablation regarding learning from fake/real data, ablation on number of input images $N$, ablation on number of directions $K$ and ablation on timesteps that edit is applied. 

\subsection{Ablations on Learning from Fake/Real Data}
In our ablation study, we explored the potential of our method to learn from synthetic images. We used single text prompts, such as \textit{a face of a person}, to generate these fake images, focusing on the domain without specifying particular attributes. Our experiments revealed that images randomly generated by Stable Diffusion (SD) often exhibit artifacts and biases, potentially affecting learning stability. Despite this, when using 100 fake images, our method successfully identified diverse directions, such as race. Fig. \ref{fig:ablation} (rightmost column) compares Gender direction learned from real (left) and fake (right) images. Both models effectively performed disentangled edits. However, the range of discovered directions using fake images was significantly narrower compared to real images, in particular we were only able to discover \textit{age, race (asian), race (indian), gender, mustache, chin shape, child} and \textit{cartoon} directions.  This limitation could be attributed to Stable Diffusion's tendency to produce flawed images with issues like crooked teeth or other artifacts, which can obstruct the learning process.

\subsection{Ablations on $N$}

Our method requires only a small set of images to learn domain-specific directions. We have found that $N=100$ images are generally sufficient for learning a rich and diverse range of directions. To explore the impact of the number of images on the discovery of directions, we conducted an ablation study using $N=10, 100, 1000$ images randomly selected from the FFHQ \cite{karras2019style} dataset, aiming to learn face-specific edits while keeping the number of directions $K=100$ constant. Our findings indicate that our method can still learn directions with as few as $N=10$ images, but the resulting directions often perform more coarse-grained edits, as shown in Race edit in Fig. \ref{fig:ablation} (first column). We believe this is due to the limited number of samples ($N=10$) available for learning our contrastive loss, providing too few positive and negative pairs to effectively learn $K=100$ directions.  Conversely, when comparing $N=100$ and $N=1000$ samples, our method demonstrates the ability to learn the same directions in both cases. This indicates that a sample size of $N=100$ is sufficient for effectively learning directions.

\subsection{Ablations on $K$}
\begin{figure*}
    \centering
    \includegraphics[width=\linewidth]{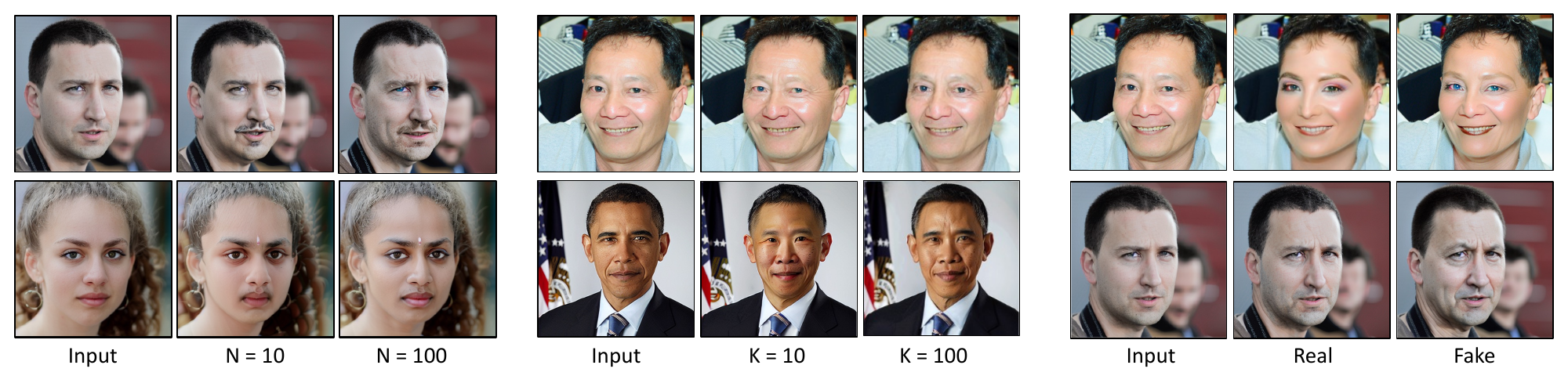}
    \caption{\textbf{Ablation study results.} We perform our ablations on three different variables, which are the usage of real/fake samples, different number of directions($K$) and number of samples used for training the model($N$). For each of our ablations, we demonstrate qualitative results on two different edits learned by each variant.}
    \label{fig:ablation}
\end{figure*}

Our method includes a hyperparameter, $K$, which determines the number of directions to be learned. For varied domains like faces or art, we typically set $K=100$, while for simpler domains like cats and cars, we choose $K=50$. In this section, we conducted an ablation study on the impact of the $K$ parameter in the face domain, keeping $N=100$ constant, and experimenting with $K={10, 50, 100}$. We observed that when the model is constrained to learn a smaller set of directions, such as $K=10$, it tends to focus on coarse-grained edits that edit the overall structure of the face, like race, age, overweight, or cartoon style. In contrast, increasing the number of directions to $K=50$ or $K=100$ leads to the discovery of more fine-grained edits, such as adjustments to lipstick, chin, eyebrows, etc. Fig. \ref{fig:ablation} (middle column) showcases Race edit discovered using $K=10$ and $K=100$. We also noticed that edits learned with $K=10$ directions are slightly more entangled than those learned with $K=100$. This could be due to the fact that directing the model to differentiate $K=100$ directions from each other enforces disentanglement, whereas a smaller number of directions may lead the model to learn more entangled edits.

\subsection{Ablations on timesteps}
\begin{figure*}[!t]
    \centering
    \includegraphics[width=1\linewidth]{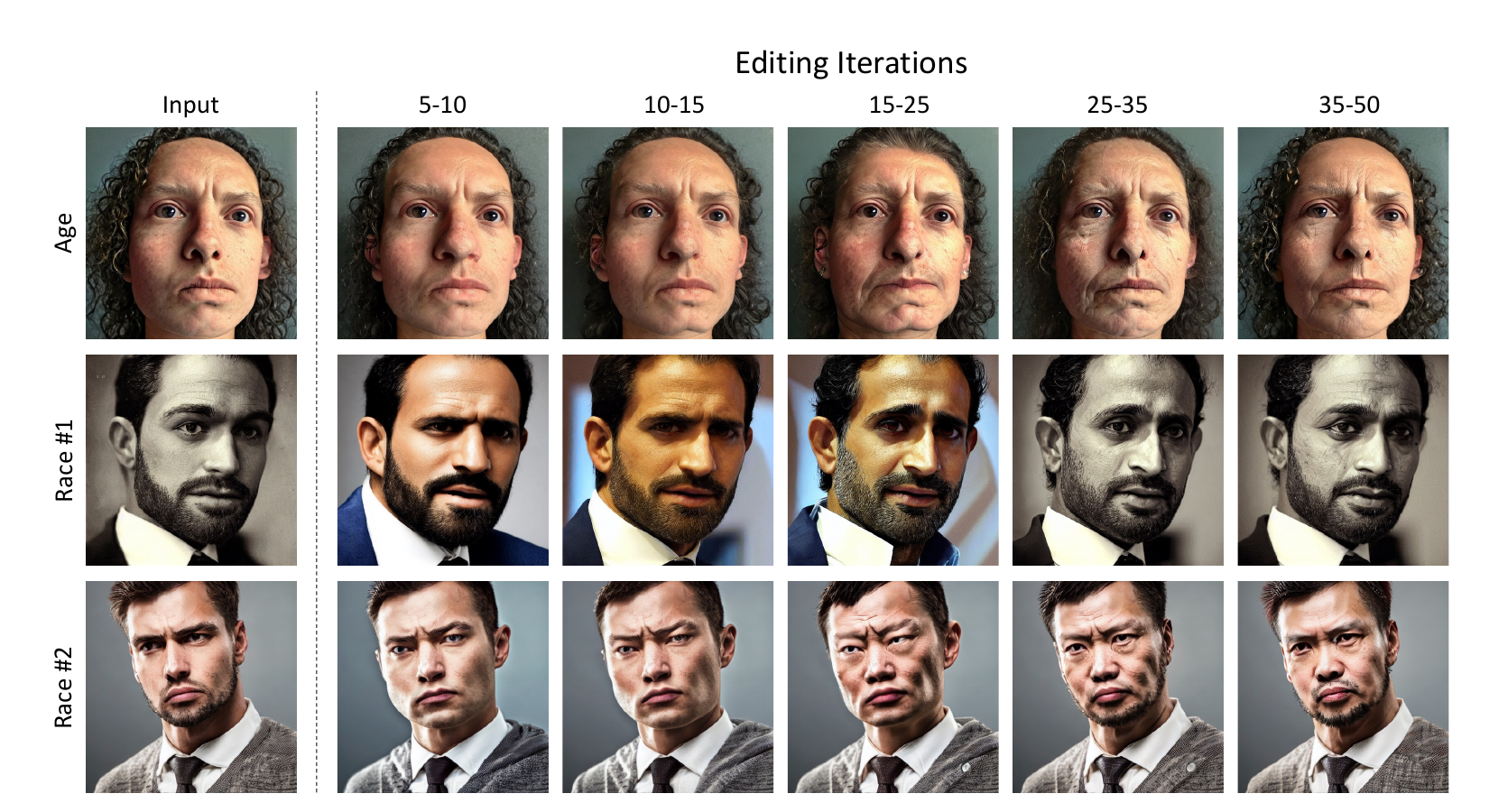}
    \caption{\textbf{NoiseCLR Ablation study on editing interval.} We demonstrate the effect of editing timesteps with ablations over age and race edits. We perform our experiments over 50 denoising steps over images generated with Stable Diffusion. For clarity, we demonstrate the applied edits on the editing iterations where iteration 0 corresponds to $t = T$ whereas iteration 50 stands for $t = 0$.}
    \label{fig:timesteps}
\end{figure*}

Prior work such as \cite{wu2023uncovering} and \cite{hertz2022prompt} has shown that timesteps are crucial factors affecting the disentanglement editing capability of Stable Diffusion. Our method, while learning directions by considering all timesteps of the diffusion model, specifically modifies the noise prediction for a certain interval of timesteps to achieve more disentangled edits. As a rule of thumb, we apply the discovered edits starting from $t = 0.5T$ to achieve disentangled edits. However, for edits that require changes in the coarse structure of the input (e.g. eyeglasses), editing at earlier timesteps are required (within the interval [$0.9T, 0.8T$]). To demonstrate the effect of timesteps on the disentanglement property of the edits, we conduct an ablation study where we apply selected edits on different timestep intervals in Fig. \ref{fig:timesteps}. In our ablations, we select the number of denoising steps as 50 and demonstrate the edited images w.r.t. denoising step indices where the edit is applied. As shown in our results, applying edits at earlier denoising steps result in more significant changes in the input image whereas edits at later iterations succeed in preserving the coarse structure of the input. 

\end{document}